\documentclass[journal]{IEEEtran}

\usepackage{amsmath}
\usepackage{epsfig}
\usepackage{graphicx}
\usepackage{amssymb,subfigure,cases}
\usepackage{setspace}
\usepackage{algorithmicx}
\usepackage{color,hyperref}
\usepackage{algorithm}
\usepackage{multirow}
\usepackage{algpseudocode}
\usepackage{array}
\definecolor{myblue}{rgb}{.73,.89,.94}
\definecolor{myorange}{rgb}{.98,.76,.39}
\definecolor{mygray}{rgb}{0.83,0.87,0.9}

\begin{document}
%

\title{A General Framework for Edited Video and Raw Video Summarization}

\author{
	Xuelong~Li,~\IEEEmembership{Fellow,~IEEE,}
	Bin~Zhao,
	and Xiaoqiang~Lu,~\IEEEmembership{Senior Member,~IEEE}
	
	\thanks{This work is supported by the National Natural Science Foundation of China (Grant No. 61761130079).}
	\thanks{Xuelong Li is with the Center for OPTical IMagery Analysis and Learning (OPTIMAL), State Key Laboratory of Transient Optics and Photonics, Xi'an Institute of Optics and Precision Mechanics, Chinese Academy of Sciences, Xi'an 710119, Shaanxi, P. R. China (xuelong\_li@opt.ac.cn).
	}
	\thanks{Bin Zhao is with The Center for OPTical IMagery
		Analysis and Learning (OPTIMAL), Northwestern Polytechnical University, Xi'an 710072, Shaanxi, P. R. China (binzhao111@gmail.com).
	}
	
	\thanks{Xiaoqiang Lu is with the Center for OPTical IMagery Analysis and Learning (OPTIMAL), State Key Laboratory of Transient Optics and Photonics, Xi'an Institute of Optics and Precision Mechanics, Chinese Academy of Sciences, Xi'an 710119, Shaanxi, P. R. China (luxq666666@gmail.com).}
		
	\thanks{\copyright 20XX IEEE. Personal use of this material is permitted. Permission from IEEE must be obtained for all other uses, in any current or future media, including reprinting/republishing this material for advertising or promotional	purposes, creating new collective works, for resale or redistribution to servers or lists, or reuse of any copyrighted component of this work in other works.}
	}

\maketitle
\begin{abstract}

In this paper, we build a general summarization framework for both of edited video and raw video summarization. Overall, our work can be divided into three folds: 1) Four models are designed to capture the properties of video summaries, i.e., containing important people and objects (importance), representative to the video content (representativeness), no similar key-shots (diversity) and smoothness of the storyline (storyness). Specifically, these models are applicable to both edited videos and raw videos. 2) A comprehensive score function is built with the weighted combination of the aforementioned four models. Note that the weights of the four models in the score function, denoted as property-weight, are learned in a supervised manner. Besides, the property-weights are learned for edited videos and raw videos, respectively. 3) The training set is constructed with both edited videos and raw videos in order to make up the lack of training data. Particularly, each training video is equipped with a pair of mixing-coefficients which can reduce the structure mess in the training set caused by the rough mixture. We test our framework on three datasets, including edited videos, short raw videos and long raw videos. Experimental results have verified the effectiveness of the proposed framework. 
	
\end{abstract}

\begin{IEEEkeywords}
	video summary, score function, property-weight, mixing-coefficient.
\end{IEEEkeywords}

\IEEEpeerreviewmaketitle

\section{Introduction} \label{section1}
\IEEEPARstart{N}{owadays}, with the popularity of camera devices, a sheer amount of videos are captured and shared online \cite{DBLP:journals/tip/TangWS16,DBLP:journals/mta/LiML16}. Every day, vast video data floods the Internet social networking platform. It provides users a convenient way to access to video data. However, it also makes data browsing time-consuming \cite{DBLP:journals/tip/LuZL17}. So we urgently need an efficient way to handle these huge video data.

Fortunately, video summary can be a good assistance in the data explosion era \cite{zhang2016context}. It offers viewers the video gist by generating a compact version of the video content. There are mainly two categories of video summary. One is storyboard, which consists of key-frames \cite{DBLP:journals/isci/HanLSHHGHL14}. The other is video skim, which is composed of video segments, namely key-shots. Usually, video shots are generated by uniform cutting or segmentation models. The two versions of video summary have individual advantages, e.g., storyboard represents the video with just a few frames, while video skim can retain the dynamic characteristics of the original video. Both of them can not only provide a viewer-friendly way to video browsing, but also have a wide range of applications, such as activity recognition \cite{DBLP:journals/tip/QinLZWS16,DBLP:journals/tii/TaoJWL14}, event detection \cite{DBLP:journals/chinaf/ZhangCLXZ16, DBLP:journals/pr/YuanWW16}, saliency detection \cite{DBLP:journals/tnn/ZhangHHS16,wang2015robust}, video embedding \cite{DBLP:journals/tcsv/ZhenSTL13,wang2017locality}, etc..

In the early years, researchers focused on the summarization of edited videos which have been preprocessed by editors, such as news, TV program, video ads and so on. Edited videos are usually with compact structures, and most shots are informative. To summarize this kind of videos, researchers focus on developing models to exploit video structure information （(i.e., the relationship among shots) \cite{DBLP:conf/iccv/NgoMZ03}, and then select the most representative elements. Most of these models are based on low-level appearance and motion features \cite{DBLP:conf/cvpr/Rav-AchaPP06, DBLP:conf/cvpr/ElhamifarSV12}.

Recently, with the popularity of camera devices, more and more videos are captured. Generally, they are formed with continuous shots and without any edit. So these videos, denoted as raw videos, are usually full of redundant and useless content. Most approaches for raw video summarization are object-based, paying more attention to describing video content (i.e., what is in the shot) \cite{DBLP:journals/pami/LiuHC10, DBLP:journals/ijcv/LeeG15, DBLP:conf/eccv/GygliGRG14}. Usually, the shots containing important objects are selected into the summary.

In this paper, we seek to build a general framework by exploiting the mutual benefit between edited video and raw video summarization. Practically, it is a challenging problem, since edited videos and raw videos have different structures. However, despite of the differences between edited videos and raw videos, they also share some commonalities in summarization. Ideally, for both edited and raw video summary, they are supposed to contain important objects (importance) and representative shots of the video content (representativeness), meanwhile with less redundancy (diversity). Last but not least, the storyline of the summary should be smooth enough to make the viewer understand the video content easily (storyness). 

More recently, score functions developed with the weighted combination of two or three of the above four properties have been employed in raw video summarization \cite{lu2013story, DBLP:conf/cvpr/LeeGG12}. However, most of their property models are built for specific kind of videos, e.g., the importance model and sotryness model in \cite{lu2013story} are designed for egocentric videos (one kind of raw videos), and cannot generalize to edited videos easily. Additionally, their property-weights are usually set manually, like in \cite{lu2013story}, since supervised learning is hard to execute due to the lack of training data.

In this paper, based on the basic idea that the summaries of both edited video and raw video share similar properties, we build a general summarization framework for them. The proposed framework can be divided into the following steps:

First, to measure the properties of video summary, we design four models, i.e., importance, representativeness, diversity and storyness. It is worth mentioning that these property models consider both the characteristics of edited video and raw video, and are applicable to the summarization of the two kinds of videos. 
 
Second, to balance the influence of the four property models, a score function is built with the weighted combination of them. The weights of the property models, denoted as property-weight, are learned in a supervised manner. It is important to note that we learn respective property-weight for edited video and raw video summarization, which is more reasonable than setting a common property-weight for both of them.

Third, to augment the training data in learning the property-weight, the training set is formed by both edited videos and raw videos. Furthermore, to reduce the structure mess caused by the rough mixture, a new parameter, denoted as mixing-coefficient, is designed for the training videos. Briefly, each training video is equipped with a pair of mixing-coefficients, reflecting its relevance to edited video and raw video summarization. 


The contributions of the proposed framework are summarized as follows:

1) We build a unified framework for the summarization of both edited videos and raw videos, which considers the commonalities and differences between the two kind of videos. 

2)	We design four models to capture the properties of video summary. They are applicable to both edited video and raw video summarization. Moreover, the score function developed with the weighted combination of the four property models can measure the summary quality comprehensively.

3) We propose to construct a combined training set with both edited videos and raw videos, which can address the problem of lacking of training data. Moreover, 
 the video structure mess in the training set is reduced by the mixing-coefficient.

The rest of this paper is organized as follows. Several classic and state-of-the-art approaches for edited video and raw video summarization are discussed in Section \ref{section2}. The detailed process of our framework is introduced in Section \ref{section3}. The experimental details and results discussion are presented in Section \ref{section4}. Finally, the conclusion is drawn in Section \ref{section5}.

\section{Related Works} \label{section2}

For a better understanding of the differences between the summarization of edited videos and raw videos, the related works are roughly classified into the following two categories.

\textbf{Edited video summarization.}

After editing, most meaningless clips are cut off, so edited videos usually have compact structures. The summarization approaches of this kind of videos focus on exploiting the frames or shots that are most representative to the video content.

Clustering is a popular technique for edited video summarization. In \cite{DBLP:journals/jvcir/KuanarPC13,DBLP:conf/icip/ZhuangRHM98,DBLP:conf/iccv/NgoMZ03,DBLP:conf/eccv/AnerK02}, the frames or shots that contain similar content are grouped into the same cluster. Generally, cluster centers are viewed as the most representative elements of the video. Then, they are selected into the summary. In earlier time, researchers \cite{DBLP:conf/sac/HadiET06,DBLP:conf/icip/ZhuangRHM98} apply classic clustering algorithms (\(k\)-means, spectral clustering, etc.) to video summarization in a simple manner. Then, more works \cite{DBLP:journals/prl/AvilaLLA11,DBLP:journals/jodl/MundurRY06} tend to modify classic clustering algorithms with the domain knowledge of video data. As in \cite{DBLP:journals/prl/AvilaLLA11}, the clusters are initialized in sequential order, since consecutive frames are usually similar and more likely to be assigned in the same cluster. Other works build more complex models to cluster the frames. In \cite{DBLP:conf/eccv/AnerK02}, the video is represented as a undirected graph, then a temporal graph is formed by partitioning it into clusters, and the summary is generated in the temporal graph. Besides, clustering is combined with other techniques to generate a summary, such as scene recognition in \cite{DBLP:conf/iccv/NgoMZ03}. Recently, a co-clusters approach is proposed in \cite{DBLP:conf/cvpr/ChuSJ15}, which summarizes several videos of the same topic simultaneously by identifying similar shots shared across these videos. 

Dictionary learning is another important technique for edited video summarization \cite{DBLP:conf/icmcs/LuanSLBLS14,DBLP:conf/cvpr/ElhamifarSV12,lu2017joint,wang2014ik}. Dictionary learning based video summarization seeks to find a few representatives (i.e., a subset of video frames or shots) to form the summary, which can be viewed as the dictionary elements.  As in \cite{DBLP:conf/icmcs/LuanSLBLS14}, it's assumed that each frame can be expressed as a linear combination of the key frames in the summary. In this case, the summary is selected by sparse coding. Furthermore, considering the local correlation of video frames, \cite{DBLP:conf/cvpr/ElhamifarSV12} employs \emph{Locality-constrained Linear Coding} (LLC) to preserve the locality. In addition, the \emph{sequential determinantal point process} (seqDPP) model is proposed to constrain the diversity of the dictionary elements \cite{DBLP:conf/nips/GongCGS14}.

\textbf{Raw video summarization.}
\begin{figure*}[t]
	\centering
	\includegraphics[width=\textwidth]{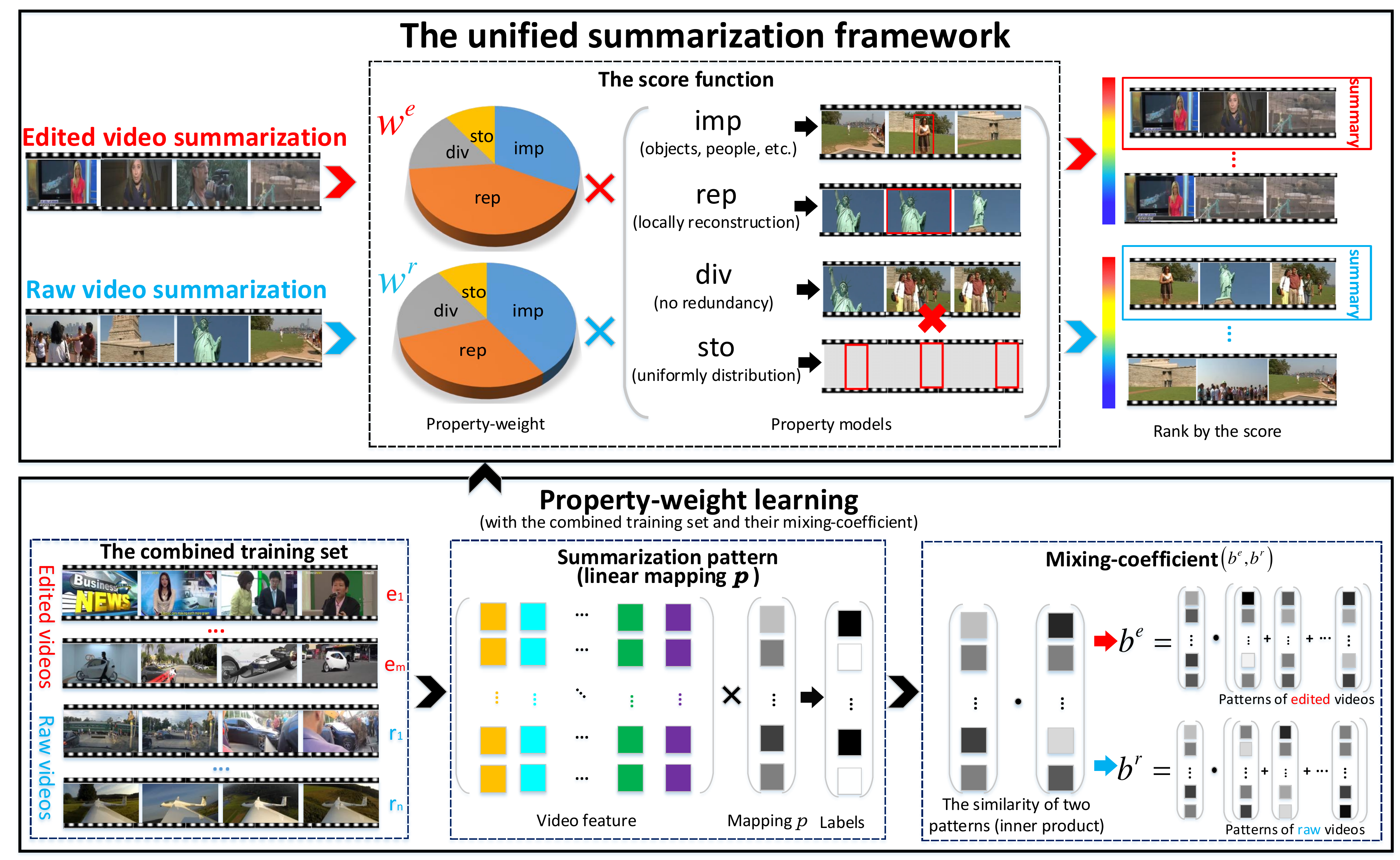}
	\caption{The detailed introduction of our work. The upper is the proposed general framework for both edited video and raw video summarization. It is a score function built with the weighted sum of four property models, i.e., importance, representativeness, diversity, storyness. (The shots displayed in the property model are all from the raw video as examples.). Note that the property-weights (i.e., \(w^e\) for edited videos, \(w^r\) for raw videos) are learned in the lower box, utilizing the combined training set equipped with mixing-coefficient \(\left( {{b^e},{b^r}} \right)\). Best viewed in color.}\label{Fig. 1}
\end{figure*}

With the popularity of cameras, the research about raw video summarization has gaining increasing attention \cite{DBLP:conf/cvpr/LeeGG12,lu2013story,DBLP:conf/eccv/GygliGRG14,DBLP:conf/eccv/PotapovDHS14}. Since there are so many meaningless shots in raw videos, a majority of the raw video summarization approaches are object-based, which focus on removing those frames or shots containing no significant people or important objects. In \cite{DBLP:journals/pami/LiuHC10}, by utilizing multi-level features, a hierarchical visual model is built to detect the desired objects.  In \cite{DBLP:conf/eccv/LiuHC08}, the irrelevant frames are removed from the video if no relevant objects are detected in them.

Recently, score functions are designed to rank the frames or shots, and the frames or shots with higher scores are selected into the summary. In \cite{DBLP:conf/cvpr/LeeGG12}, the summarization is formulated as scoring the video frames according to their visual importance, which is modeled by several features describing the presence of faces, objects, etc.. Similar with \cite{DBLP:conf/cvpr/LeeGG12}, an interesting model is built in \cite{DBLP:conf/eccv/GygliGRG14} by combining the features that describe the aesthetics of frames, the presence of landmarks and so on. Moreover, in \cite{lu2013story}, multi-models are employed to avoid the redundancy and gaps in understanding the summary. Specifically, these models are developed for egocentric videos, and due to the lack of training data, the weights of these models are set manually rather than learned in a supervised manner. More recently, the convolutional neural network is employed to score the video shots in \cite{Yao_2016_CVPR}.

\section{The General Summarization Framework} \label{section3}

The flow chart about the whole proposed framework is depicted in Fig. \ref{Fig. 1}, and the details about each part are introduced in the following subsections.

\subsection{Property models}\label{section3.2}

\textbf{Importance.}

Some of the elements appearing in the video are quite important for viewers to understand the video content, such as the landmarks in travel videos, the products in television ads, the man you are talking to in egocentric videos. The frames containing these elements should be well-protected in the summarization procedure, since they contain lots of important information.

In this paper, the regression model proposed in \cite{DBLP:conf/eccv/GygliGRG14} is employed to measure the importance. It is originally proposed for raw videos, which predicts the importance of the video frame with three kinds of features. The first kind of features denote the aesthetic and quality of video frames. The second kind of features represent the likelihood of faces, persons, landmarks in the frame, which are calculated via several detectors. The third kind of features describe the probability of an object to be a follow-object (i.e., tracked by the camera). 

Unfortunately, the third kind of features are not well suited to edited videos. Because the edited video is composed of several short shots, there are few follow-objects. Fortunately, the probability determined by the third kind of features can be regarded as the importance of the detected object. In this case, we try to replace the third kind of features with others that can achieve this goal and are applicable to edited videos. In view of that the second kind of features in \cite{DBLP:conf/eccv/GygliGRG14} employ several detectors, we believe that a large detected bounding box in the middle of the frame indicates higher importance of the detected object than a small bounding box in the corner of the frame. So the sizes and locations of the bounding boxes are utilized to replace the third kind of features. Then, the importance score of each frame is calculated by the regression model in \cite{DBLP:conf/eccv/GygliGRG14}.

In our model, the importance of the shot is the sum over the importance of its frames. Moreover, the total importance of the key shots are denoted as the importance of the summary, which is described as follows:

\begin{equation}{f^{imp}}\left( {V,S} \right) = \frac{1}{{{M^{imp}}}}\sum\limits_{s \in S} {I\left( s \right)}, \end{equation}
where \(V\) and \(S\) stand for the feature matrix of video and its candidate summary, respectively. \(I\left( s \right)\) is the importance of the shot \(s\). \({f^{imp}}\left( {V,S} \right)\)  reflects the ability of \(S\)  to capture the important elements in the video \(V\), whose range is normalized to \([0,1]\) by the total importance of the video frames, \({{M^{imp}}}\).

\textbf{Representativeness.}

In fact, the summary should reflect the main content of the original video, denoted as the representativeness. However, the summary is just a subset of the video shot and the information loss of the summarization is inevitable. So it is necessary to build the representativeness model to constrain the information loss.

In our model, the representativeness stands for the ability of the summary to cover the video information, which is measured by the reconstruction error (i.e., reconstructing the video with its summary). Specifically, due to the continuity of the video storyline, the video shots are locally related. Therefore, different from traditional models \cite{DBLP:conf/cvpr/ZhaoX14a, DBLP:conf/icmcs/LuanSLBLS14} that reconstruct the video globally, we reconstruct each video shot with its near key shots, and utilize the following function to calculate the reconstruction error:
\begin{equation}E\left( {x,S} \right) = \min \left( {{{\left\| {x - {l_x}} \right\|}_2},{{\left\| {x - {r_x}} \right\|}_2}} \right), {l_x},{r_x} \in S,\;\forall x \in V,\end{equation}
where \({l_x}\)  and \({r_x}\)  are the nearest key-shots on the left side and right side of the shot \(x\), respectively. The intuition lying behind Equ. (2) is that if the Euclidean distance between shot \(x\) and its nearest key shot is small, the content in \(x\)  is captured by the summary. 

Then, the representativeness of the summary can be quantified as:
\begin{equation}{f^{rep}}\left( {V,S} \right) = \frac{1}{{{M^{rep}}}}\sum\limits_{x \in V} {\exp \left\{ { - E\left( {x,S} \right)} \right\}}, \end{equation}
where \({{M^{rep}}}\) is the maximal value of representativeness. It achieves when all shots are selected into the summary.

\textbf{Diversity.}

 The property of diversity is to constrain that there is no redundancy, which is necessary for the simplicity of video summary.
 
 Previous works have built various models to measure the diversity of the summary \cite{DBLP:conf/icip/AnirudhMT16, DBLP:journals/tmm/ShroffTC10, DBLP:conf/nips/GongCGS14}. While for the efficiency of our framework, we try to build a model that are not only able to reduce the redundancy but also computationally efficient. Based on this, the following function is developed:
\begin{equation}{f^{div}}\left( {V,S} \right) = \frac{1}{{{M^{div}}}}\sum\limits_j {{{\left\| {{s_j} - {s_{j + 1}}} \right\|}_2},\quad } {s_j},\;{s_{j + 1}} \in S,\end{equation}
where \({s_j}\)  and \({s_{j+1}}\) are consecutive shots in summary \(S\). \({{M^{div}}}\) is a constant used to normalize the range of the diversity from 0 to 1. Considering that \({f^{div}}\) is an increasing function on the number of shots in the summary \(S\), \({{M^{div}}}\) is fixed as the total \({\left\|  \cdot  \right\|_2}\)  differences of consecutive key shots when all video shots are selected into \(S\). 

Equ. (4) constrains that consecutive key shots should be different from each other, while non-consecutive ones needn't. It is done for two reasons: 1) Consecutive shots are more likely to be similar with each other, while the non-consecutive shots not. Usually, the summary meets the demand of diversity if consecutive key shots are dissimilar. 2) The non-consecutive shots may offer different information if they are similar, e.g., the two non-consecutive shots about two goals in a soccer video. Although they are similar, both of them should be preserved in the summary.

\textbf{Storyness.}

The property of storyness demands that the information displayed in the summary should not only reflect the main video content, but also be easy to understand. The concept of storyness is first proposed for egocentric video summary in \cite{lu2013story}, which assumes that a summary tells a good story if the objects in two consecutive shots are strong related. Unfortunately, in edited videos, relationships between consecutive shots have been broken by manual editing. For example, the shots switch between the studio and spot in a news video. Therefore, the relationship between visual objects in consecutive shots of the edited video is not as strong as raw videos. That's to say, the storyness model in \cite{lu2013story} is not applicable to edited videos. Therefore, we tend to build a more general storyness model.

Actually, for both edited and raw videos, a summary is of high storyness if there are no gaps in the storyline. To achieve this, the key shots in the summary are expected to distribute uniformly. So the storyness function can be described as follows:
\begin{equation}{f^{sto}} = \exp \left\{ { - \sum\limits_j {\left| {{L_j} - \frac{{\left| V \right|}}{{\left| S \right|}}} \right|} } \right\},\end{equation}
where \begin{equation}{L_j} = Index\left( {{s_{j{\rm{ + }}1}}} \right){\rm{ - }}Index\left( {{s_j}} \right),\quad {s_j},{s_{j + 1}} \in S.\end{equation}
\(Index\left( {{s_j}} \right)\) is the index of \({s_j}\) in the original video. \(\left| V \right|\) and \(\left| S \right|\) are the number of shots in the original video and the summary, respectively. According to Equ (6), if the key shots are scattered from the beginning to the end of the video, the mean interval between consecutive key shots is \({\frac{{\left| V \right|}}{{\left| S \right|}}}\) approximately. The exponential form of Equ. (5) is just used to normalize the value from 0 to 1. From Equ. (5), we can see that if the key shots are distributed uniformly in the video, the summary achieves the highest storyness score. The intuition lying behind this is that the uniform distribution indicates the smoothness of the storyline in the summary, which means the summary content is easy to understand.

\subsection{Score Function}\label{section3.1}

According to the definition of the four properties, it can be seen that they measure the quality of the summary in different aspects, and constrain the summary to be a good epitome of the original video. But they may cause interferences in the summarization procedure if not dealt well with. For example, if the importance property is emphasized too much, the summary may be full of shots containing specified important objects. But the other properties are destroyed. In view of this, we build a comprehensive score function to balance the influence of these properties.

As we know, the video summary is a subset of the video shots. Essentially, the summarization procedure is to find the subset that can best summarize the video content. To generate the wanted summary, we develop a score function to rank the subset of the video shots:
\begin{equation}{S^*} = \arg \mathop {\max }\limits_{S \subset V} F\left( {V,S,w} \right),\end{equation}
where \(F\) is the score function that is used to exploit the hidden correlation between the video and its summary. \(w\) is a parameter, introduced detailedly in the following. In this function, the subset of the video shots \(S\) is mapped to a score which reflects its capability to summarize the video content. Not surprisingly, the desired summary \({S^*}\) is pre-defined as the one that can achieve the highest score. 

 In this paper, the score function is built by the linear combination of the aforementioned property models, which is described as follows:
\begin{equation}F\left( {V,S,w} \right) = {w^{\rm T}}f\left( {V,S} \right),\end{equation}
where  \(f = {\left\{ {{f^{imp}},{f^{rep}},{f^{div}},{f^{sto}}} \right\}^{\rm T}}\), \(w\) is the property-weight, which represents different emphases of the summarization on the four summary properties. 

In our framework, \(w\) is learned in a supervised manner. In the learning procedure, videos and their human-made summaries are taken as the training data. However, for raw videos, since the theme is not obvious and shots transform smoothly, people share less agreement on the wanted-summary \cite{DBLP:conf/eccv/GygliGRG14}. So the training data is hard to collect for raw videos. To address this problem, the training set is formed with the combination of both edited videos and raw videos. Actually, we want edited videos help raw videos in the training procedure.

Till now, there are still two problems need to be settled:

1)	The training set. Due to the differences of the structures between edited videos and raw videos, their summarization patterns may vary largely from each other. So the rough combination of edited videos and raw videos may cause interference in the training procedure, even outweighs their contributions to extend the training set.

2)	the property-weight. Although both edited video and raw video summaries share the aforementioned four properties, they have individual characteristics in summarization, embodied in their different emphases on these properties. For example, raw videos put more emphases on the importance property, because they have many shots with no significant objects. While edited videos not, since most shots (left after editing) are full of these important objects. They focus more on finding shots which are most representative to the video content. So it is not reasonable to summarize edited videos and raw videos indiscriminately with the same property-weight. Ideally, we want to develop individual property-weight for edited videos and raw videos.

In the following subsections, both of the two problems are addressed by introducing a new parameter to the combined training set, i.e., mixing-coefficient.

\subsection{Mixing-coefficient}\label{section3.3}
\begin{figure}[t]
	\centering
	\includegraphics[width=0.5\textwidth]{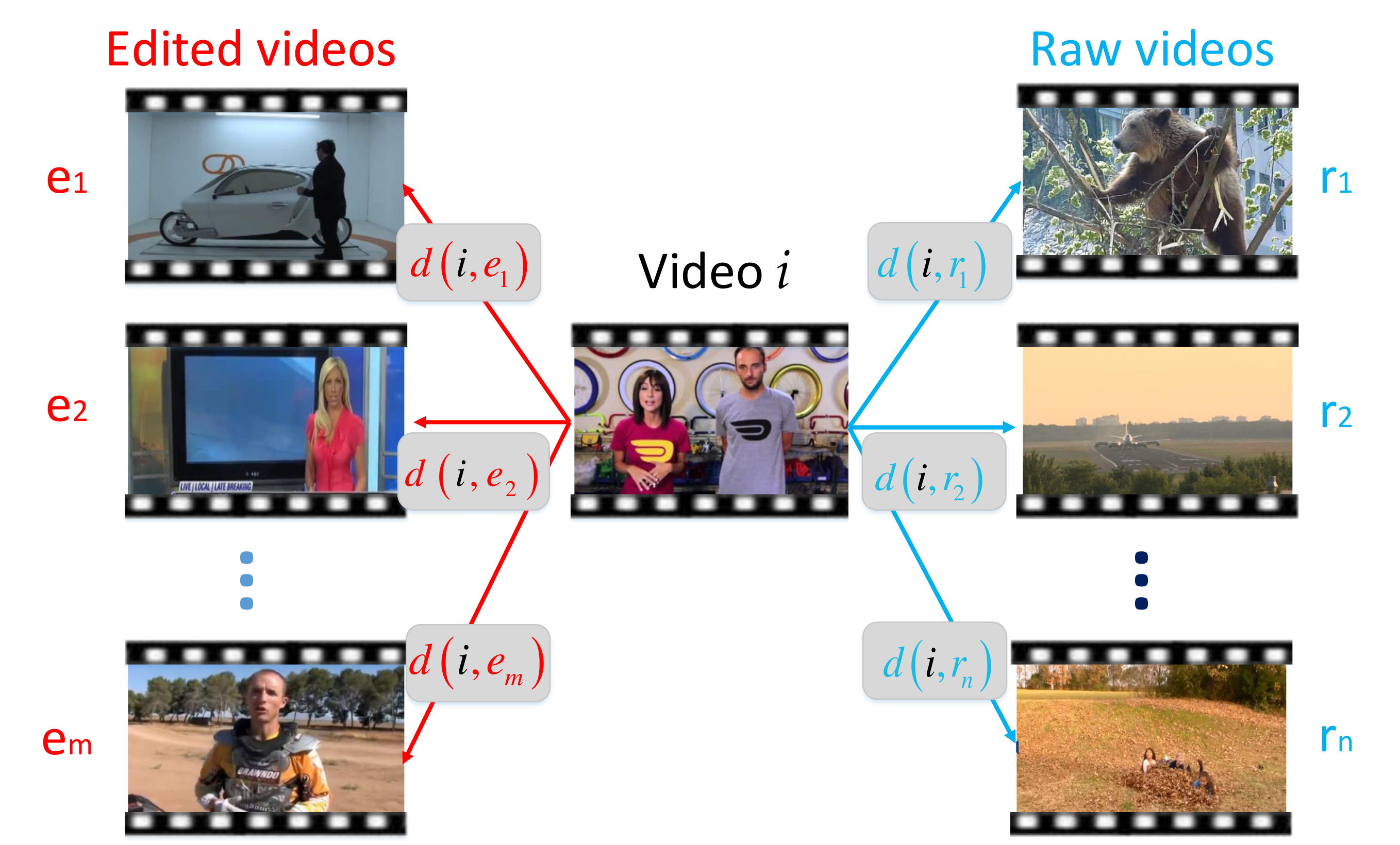}
	\caption{The distribution of summarization pattern similarity \(d\) between video \(i\) and all the other videos in the training set. Where \(d\left( {i,e} \right)\) denotes the similarity between the summarization patterns of video \(i\) and the edited video \(e\). The mean of \(d\left( {i,e} \right),e \in edited\) determines the mixing-coefficient \({b_i^e}\). Similarly, the mean of \(d\left( {i,r} \right),r \in raw\) determines the mixing-coefficient \({b_i^r}\).}\label{b_e_and_b_r}
\end{figure}

We design a pair of mixing-coefficient \(\left( {{b^e},{b^r}} \right)\) for each training video when mixed to the combined training set. Actually, \({{b^e}}\) and \({{b^r}}\) stand for the relevance of a certain video to edited and raw video summarization, respectively. The details about the calculation of \({{b^e}}\) and \({{b^r}}\) are described as follows.

 Firstly, according to \cite{DBLP:conf/cvpr/ElhamifarSV12,DBLP:conf/icmcs/LuanSLBLS14}, we employ the following model to calculate the mapping of the video to its summary:
\begin{equation}p = \mathop {\arg \min }\limits_p {\left\| {{V^{\rm T}}p - y} \right\|_2} + {\left\| p \right\|_1},\end{equation}
where \(V \in {R^{k \times q}}\) denotes the feature matrix, \(q\) is the number of shots in the video and \(k\) is the dimensionality of the shot features. \(y \in {R^q}\) is the label of the video shot (i.e., 1 for key shot, 0 for others). \(p\) is vector which encodes the mapping from the video to its summary. In this paper, \(p\) is denoted as the summarization pattern of the video. \({\left\|  \cdot  \right\|_1}\) constraints \(p\) to be sparse.

Then, the following equation is used to describe the similarity of two patterns:
\begin{equation}d\left( {i,j} \right) = \exp \left\{ {\frac{{\left( {{p_i},{p_j}} \right)}}{{\left| {{p_i}} \right| \cdot \left| {{p_j}} \right|}}} \right\}.\end{equation}
It can be seen in Equ. (10) that the similarity of the two summarization patterns is determined by the inner product of their mapping vector directions (i.e., normalized vector). The closer directions of the mapping vectors imply more similar summarization patterns of two videos. In the training set, the similarities of summarization patterns (i.e., \(d\)) between a certain video and all the other videos is depicted in Fig. \ref{b_e_and_b_r}.

Furthermore, for a certain video \(i\) in the training set, its mixing-coefficient \({b_i^e}\), which reflects the relevance of video \(i\) to edited video summarization, is calculated by averaging the similarities between its summarization pattern and those of edited videos. Similarly, \(b_i^r\) is calculated by averaging the similarities between its summarization pattern and those of raw videos. Specifically, they are formulated as:

\begin{equation}b_i^e = \left\{ \begin{array}{l}
\frac{\alpha }{m}\sum\limits_{j \in edited} {d\left( {i,j} \right)} ,\quad if\;i \in raw\\
\frac{\alpha }{{m - 1}}\sum\limits_{j \in edited\backslash i} {d\left( {i,j} \right)} ,\quad if\;i \in edited
\end{array} \right.,\end{equation}
\begin{equation}b_i^r = \left\{ \begin{array}{l}
\frac{\beta }{n}\sum\limits_{j \in raw} {d\left( {i,j} \right),\quad if\;i \in edited} \\
\frac{\beta }{{n - 1}}\sum\limits_{j \in raw\backslash i} {d\left( {i,j} \right),\quad if\;i \in raw} 
\end{array} \right.,\end{equation}
where \(m\) and \(n\) are the number of edited videos and raw videos in the training set, respectively. \(\alpha \) and \(\beta\) are fixed constants that constrain \(\max \left\{ {{b^r}} \right\} = 1\) and \(\max \left\{ {{b^e}} \right\} = 1\). 

Finally, each video \(i\) in the training set is equipped with a pair of mixing-coefficient \(\left( {b_i^e,b_i^r} \right)\). Practically, the training videos together with their mixing-coefficients are utilized to learn the summarization property-weights of the two kinds of videos. 

\subsection{Property-weight}\label{section3.4}

In our approach, the property-weight is learned individually for edited videos and raw videos. Specifically, edited videos share the weight vector \({w^e}\), while raw videos share the other weight vector \({w^r}\). 

In this paper, the maximum margin approach \cite{DBLP:conf/uai/LinB12} is employed to learn the property-weight, where the ground truth of the summary is expected to achieve a higher score than all the other summary candidates by some margin. The structured hinge loss function is suitable for this problem \cite{roller2004max}:
\begin{equation}{L_V}\left( w \right) = \mathop {\max }\limits_{S \subset V} \left[ {F\left( {V,S,w} \right) + \Delta \left( {S,{S^{gt}}} \right)} \right] - F\left( {V,{S^{gt}},w} \right),\end{equation}
where \({S^{gt}}\)  is the ground truth of the video summary. \(\Delta \left( {S,{S^{gt}}} \right)\) is the task loss, which is defined as:
\begin{equation}\Delta \left( {S,{S^{gt}}} \right) = \frac{{\left| {S \cap \left( {V\backslash {S^{gt}}} \right)} \right|}}{{\left| {{S^{gt}}} \right|}}.\end{equation}
It counts the percentage of the summary shots not presented in the ground truth. According to \cite{roller2004max}, the task loss ensures that only the ground truth can get the highest score of \(F\). 

Practically, since the mixing-coefficients reflect the relevance of each training video to a specific task (i.e., edited video summarization or raw video summarization), they are also employed in this part as the training coefficient of each video. Specifically, with the mixing-coefficient \({b^e}\), the modified optimization function for learning the property-weight of edited video summarization \(w^e\) is:
\begin{equation}{w^e} = \arg \mathop {\min }\limits_{{w^e} \ge 0} \frac{1}{{\left| T \right|}}\sum\limits_{V \in T} {b_V^e} {L_V}\left( {{w^e}} \right) + \frac{\lambda }{2}{\left\| {{w^e}} \right\|^2,}\end{equation}
where \(T\) is the training set, combined with raw videos and edited videos. \(\lambda \)  is the parameter balancing the weights of the loss term and the regularization term, fixed as 0.01 in this paper. Actually, the regularization term is employed to smoothen the distribution of the elements in \(w^e\). It can be observed that Equ. (15) is convex, since its first term is a linear function of \({w^e}\), and the second term is a quadratic function of \({w^e}\). 

Similarly, \({w^r}\) is learned in the same way with mixing-coefficient \({b^r}\), formulated as:
\begin{equation}{w^r} = \arg \mathop {\min }\limits_{{w^r} \ge 0} \frac{1}{{\left| T \right|}}\sum\limits_{V \in T} {b_V^r} {L_V}\left( {{w^r}} \right) + \frac{\lambda }{2}{\left\| {{w^r}} \right\|^2.}\end{equation}


%





%

\subsection{Optimization}\label{section3.5}

\begin{algorithm}
	\label{Alg-1}
	\caption{Optimization for property-weight}
	\noindent\textbf{Input}: training set \(\left\{ {{V_n},{S^{gt}},b_n^e,b_n^r} \right\}_{n = 1}^N\), learning rate \(\left\{ {{\gamma_n}} \right\}_{n = 1}^N\), iterations \(I\), maximal length of the summary \(C\).
	\noindent\textbf{Output}: \(w^e\) and \(w^r\)
	\begin{algorithmic}[1]
		\State\textit{\textbf{1. Calculate \(w^e\) and \(w^r\).}}	
		\For {$n=1$ to $N$}
		\State${w_n} = PSD\left( {{V_n},S_n^{gt},{\gamma _n}} \right)$;
		\EndFor
		\State${w^r}{\rm{ = }}\frac{1}{B^e}\sum\limits_{n = 1}^N {b_n^e} {w_n}$, where ${B^e} = \sum\limits_{n = 1}^N {b_n^e}$;
		\State${w^e}{\rm{ = }}\frac{1}{B^r}\sum\limits_{n = 1}^N {b_n^r} {w_n}$, where ${B^r} = \sum\limits_{n = 1}^N {b_n^r}$;\\
		
		\State\textit{\textbf{2. Projected Subgradient Descent Algorithm.}}
		\Function {$PSD$} {$V,{S^{gt}},\gamma $}
		\State ${w^{\left( 0 \right)}} = 0$;
		\For {$t=1$ to $I$}
		\State ${S^{*\left( t \right)}} = AG\left( {V,{\left| {{S^{gt}}} \right|},w} \right)$;
		\State ${g^{\left( t \right)}} = \lambda {w^{\left( {t - 1} \right)}} + f\left( {V,{S^{*\left( t \right)}}} \right) - f\left( {V,{S^{gt}}} \right)$;
		\State ${w^{\left( t \right)}} = \max \left( {0,{w^{\left( {t - 1} \right)}} - \gamma {g^{\left( t \right)}}} \right)$;
		\EndFor 
		\State \Return $w = {w^{\left( I \right)}}$;
		\EndFunction
		\State\textit{\textbf{3. Accelerated Greedy Algorithm.}}
		\Function {$AG$}{$V,C,w$}
		\State $S = \emptyset $;
		
		\State \textbf{while} $\left| S \right| < C$ \textbf{do}
		\State $\hat s = \arg \mathop {\max }\limits_{s \in V\backslash S} F\left( {V,S \cup s,w} \right)$;
		\State $S = S \cup \hat s$;
		\State \textbf{end while}
		\State \Return ${S^*} = S$;
		\EndFunction
		
	\end{algorithmic}
\end{algorithm}

To optimize Equ. (15) and Equ. (16), we should first find the optimal solution \(S\) by maximizing Equ. (13) with a given \(w\).  It is an optimal subset selection problem. However, there are totally \({2^{\left| V \right|}}\) shot subsets in the video, where \(\left| V \right|\) is the number of shots. So it is time consuming to compute the score for every subset and ranking them to find the optimal solution. To address this problem, submodular function is employed in our work. 

\begin{figure}[t]
	\centering
	\includegraphics[width=0.3\textwidth]{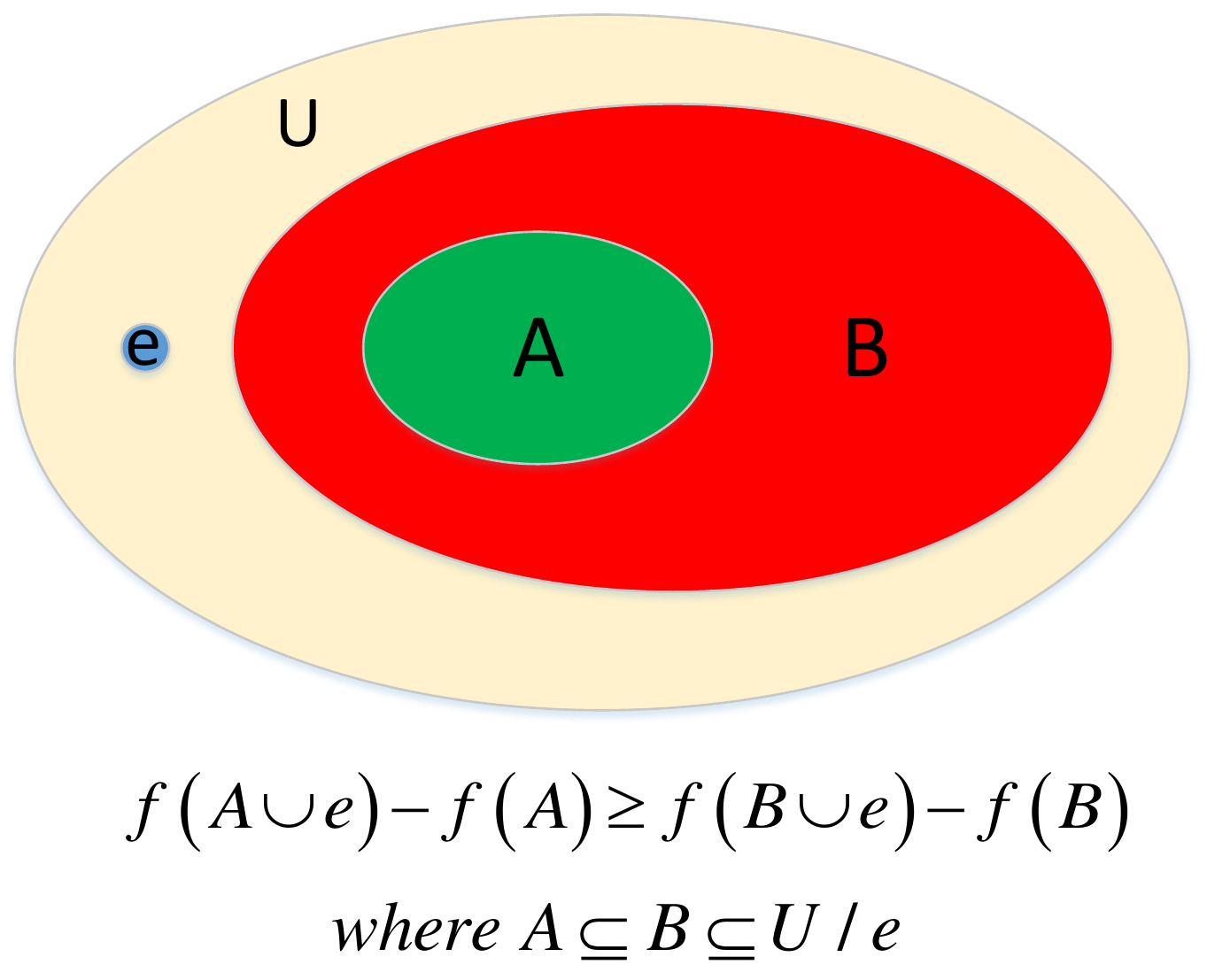}
	\caption{The diminishing gains property of submodular function. In detail, \(f\left( {A \cup e} \right) - f\left( A \right)\) is denoted as the gain of set function \(f\) by adding a new element \(e\) to \(A\). If \(f\) if submodular, for \(\forall B \supseteq A\) and \(e \notin B\), the gain of \(f\) by adding \(e\) to \(B\), i.e., \(f\left( {B \cup e} \right) - f\left( B \right)\) is less than \(f\left( {A \cup e} \right) - f\left( A \right)\). That is the meaning of diminishing gains.}\label{fig_submodular}
\end{figure}

Firstly, we provide a brief introduction to the submodular function. As introduced in \cite{fujishige2005submodular}, a set function is submodular if it satisfies the property of diminishing gains, as described in Fig. \ref{fig_submodular}. It explains that the gains of submodular function are reducing with the increasing of selected elements in the set. Submodular functions have two advantages: 1) The non-negative weighted sum of submodular functions is also submodular \cite{DBLP:conf/acl/LinB11}. 2) The optimization of submodular functions can be solved approximately by greedy algorithms with rigorous performance guarantee \cite{feige2011maximizing}. The two advantages lead to the popularity of submodular optimization in document summarization \cite{DBLP:conf/acl/LinB11, DBLP:conf/asru/LinBX09} and image collection summarization \cite{DBLP:conf/iccv/SimonSS07, tschiatschek2014learning}.

In fact, the video summarization procedure in this paper also satisfies the property of diminishing gains. This is because that we prefer to earlier select the shots that meet the four properties best, the gains diminish along with the selection going on. Inspired by this, according to \cite{gygli2015video}, the four property models are reformulated to the submodular form. Based on the first advantages of the submodular function, Equ. (8) is also submodular because its the non-negative weighted sum of submodular functions. In this case, to find the optimal subset of the video shots, the \emph{Accelerated Greedy} (AG) algorithm \cite{minoux1978accelerated} is employed to maximize Eq. (13). Finally, to determine the property-weight \(w^e\) and \(w^r\), Equ. (15) and Equ. (16) are optimized with the \emph{Projected Subgradient Descent} (PSD) algorithm, which is derived from the standard \emph{Subgradient Descent} (SD) \cite{ratliff2006subgradient} by constraining the property-weight to be non-negative. The convergence of PSD have been proved in \cite{DBLP:conf/uai/LinB12}. Concisely, the detailed optimization procedure for property-weight is depicted in Algorithm 1.

Besides, in the test procedure, the main computations of generating a video summary lie in the \emph{Accelerated Greedy} (AG) algorithm. According to \cite{minoux1978accelerated}, the computational complexity is \(O\left( {\left| V \right|} \right)\).

\section{Experimental Results} \label{section4}

\begin{figure*}[!t]
	\centering
	\subfigure[Training coefficient for edited video summarization]{
		\label{Traing_weight_for_edited}
		\includegraphics[width=0.4\textwidth]{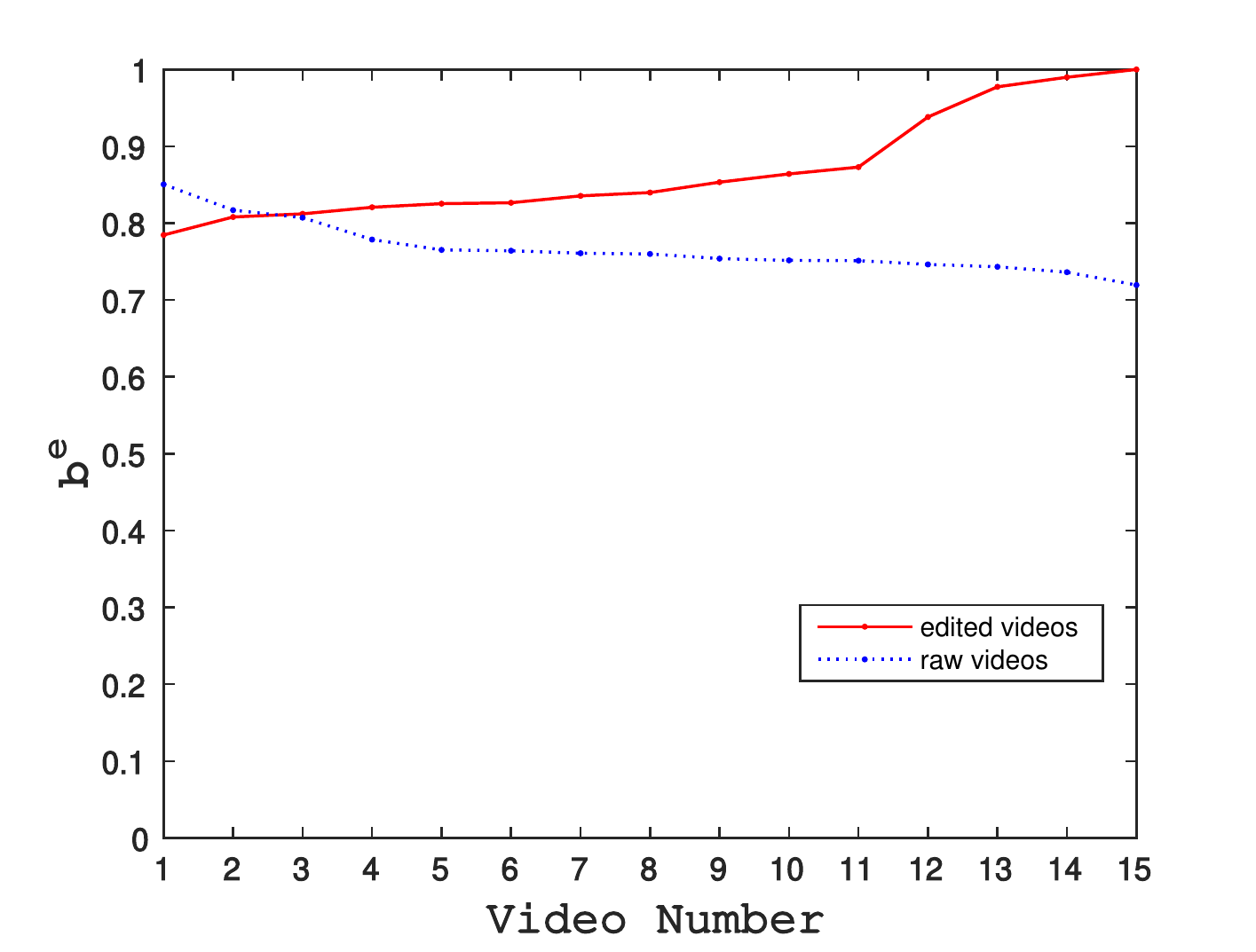}}
	\subfigure[Training coefficient for raw video summarization]{
		\label{Training_weight_for_raw}
		\includegraphics[width=0.4\textwidth]{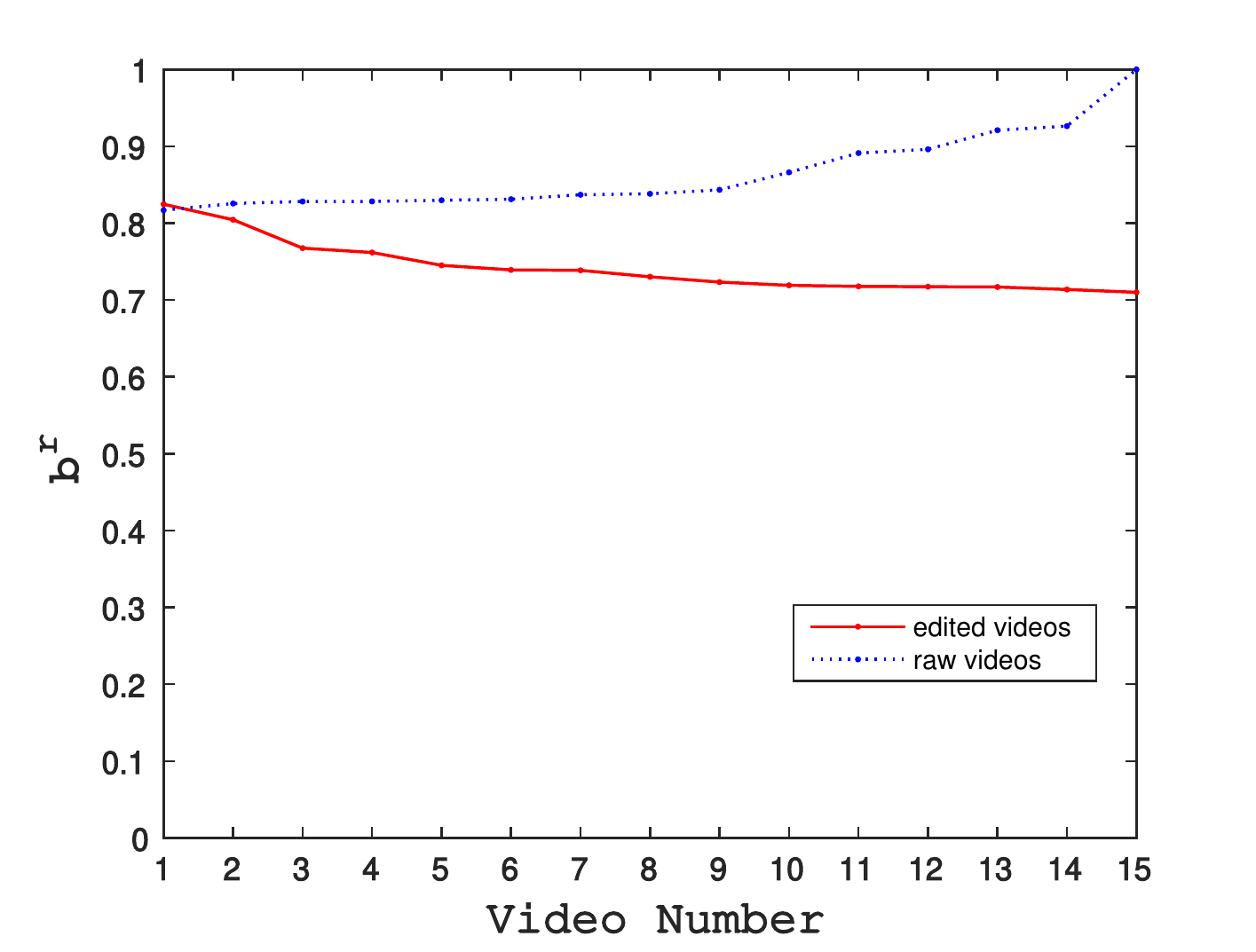}}
	\caption{The distribution of training coefficient (mixing-coefficient) in the training set. Note that, in the left sub-figure, the \(b^e\) of edited videos is ascending-sorted, while the \(b^e\) of raw videos is descending-sorted. It doesn't mean the distribution of \(b^e\) has obviously upward or downward trend, but just for easily comparing their values. The \(b^r\) is sorted for similar purpose.} \label{Training_weight}
\end{figure*}

\begin{figure*}[!t]
	\centering
	\subfigure[Property-weight for edited video summarization]{
		\label{Property_weight_for_edited}
		\includegraphics[width=0.4\textwidth]{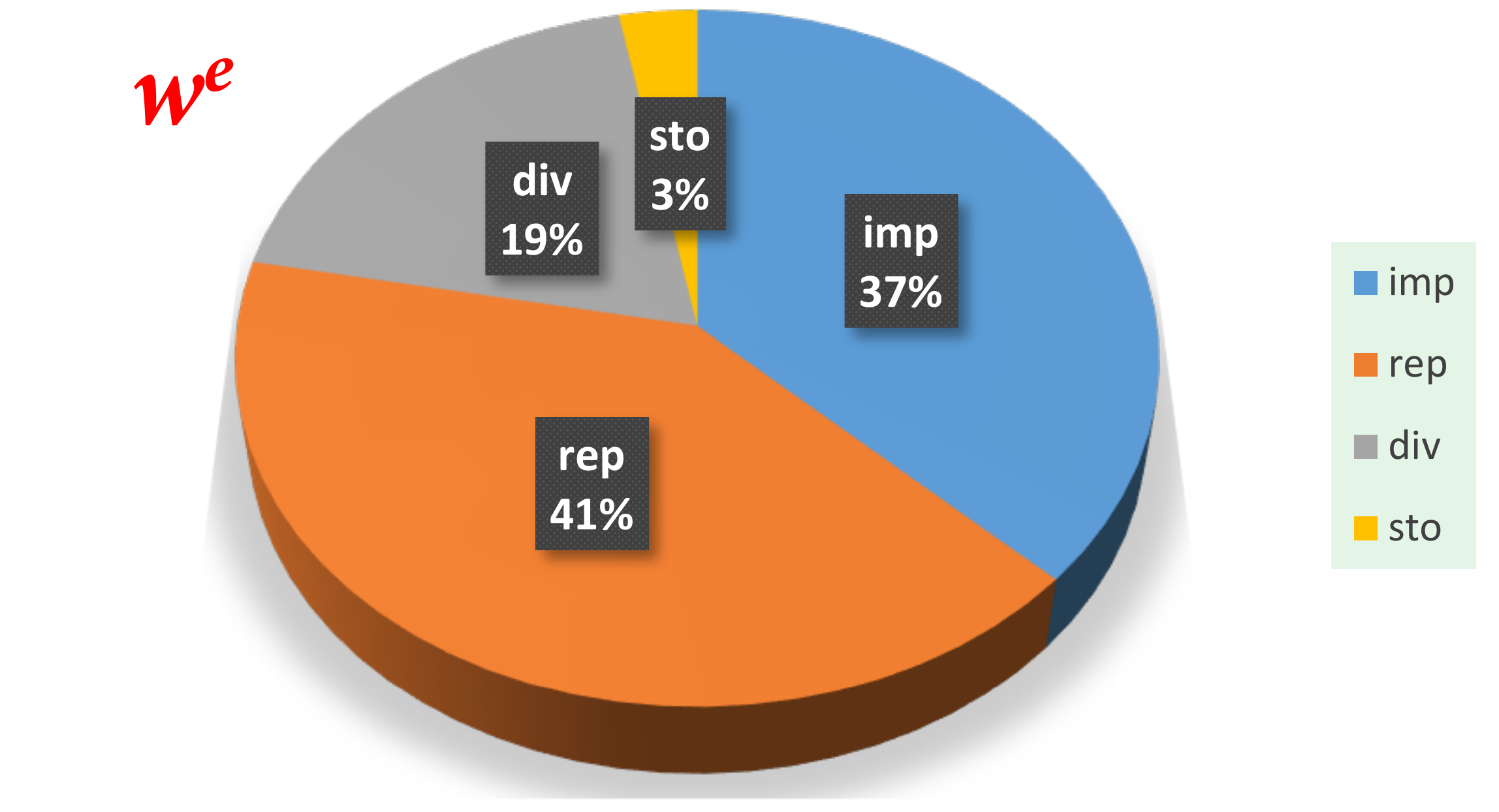}}
	\subfigure[Property-weight for raw video summarization]{
		\label{Property_weight_for_raw}
		\includegraphics[width=0.4\textwidth]{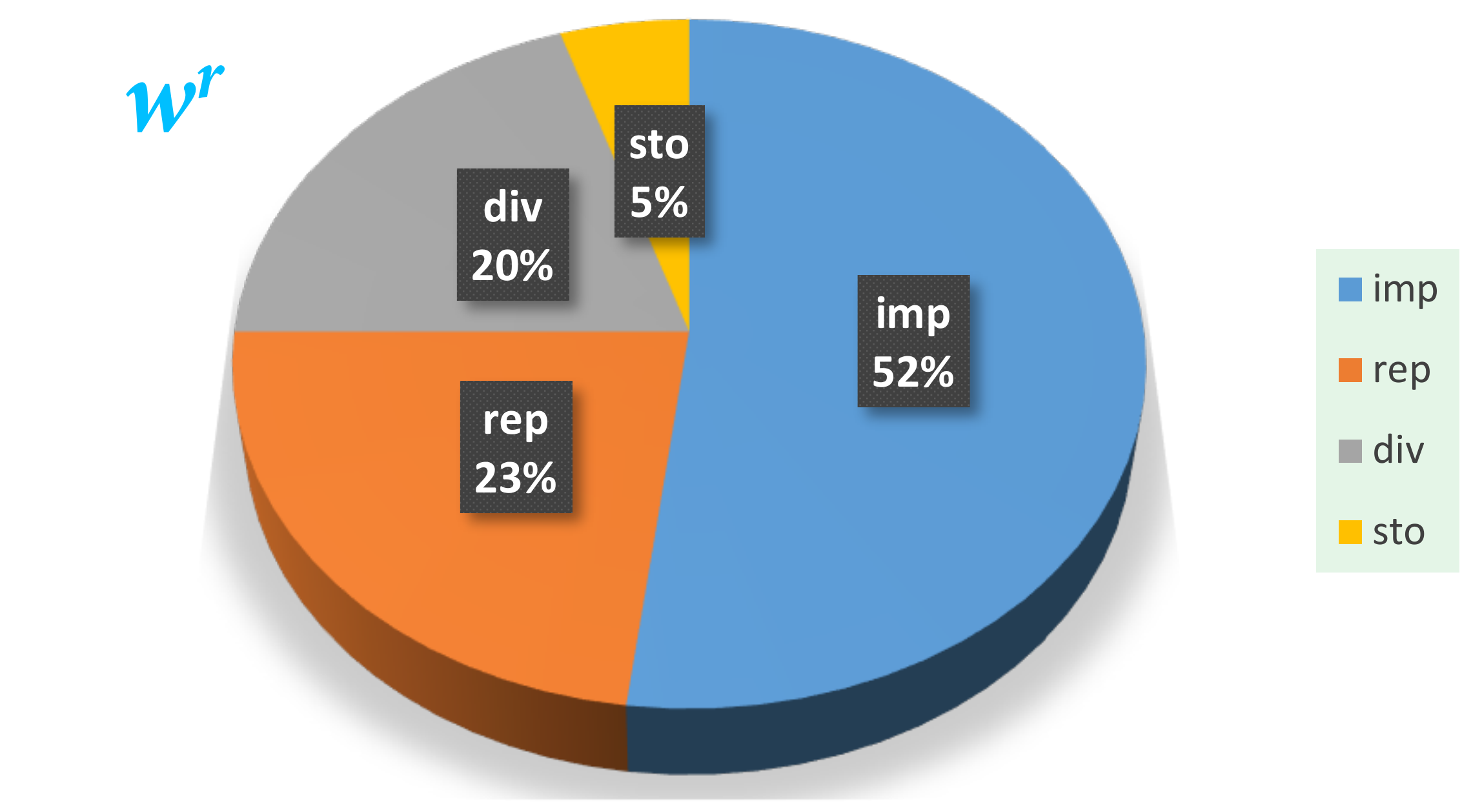}}
	\caption{The property-weight, \(w^e\) and \(w^r\), for edited and raw video summarization, respectively. } \label{Property_weight}
\end{figure*}

The proposed summarization framework is verified on 3 datasets, i.e., TVsum50 \cite{song2015tvsum}, SumMe \cite{DBLP:conf/eccv/GygliGRG14}, ADL \cite{pirsiavash2012detecting}, for edited videos, short raw videos and long raw videos, respectively. Based on these datasets, the performance of the proposed framework is compared with various popular approaches about edited video and raw video summarization.

\subsection{Implementation details}\label{section4.1}

\emph{Training set.} The training set is composed of 15 edited videos and 15 raw videos. These videos are randomly selected from the TVsum50 dataset (edited videos) and the SumMe dataset (raw videos), respectively. Each video is segmented into about 50-300 shots (More details about the datasets and shot segmentation methods are described in the dataset introduction part of Section \ref{section4.2} and \ref{section4.3}). Practically, there are more than 4000 shots for training. Each shot is annotated with user scores, and the ground truth of the summary is constructed with high-score shots.

As aforementioned, the mixing-coefficients \(\left( {{b^e},{b^r}} \right)\) of training videos are employed as their training coefficient in learning the property-weight, which are
plotted in Fig. \ref{Training_weight}. In Fig. \ref{Traing_weight_for_edited}, it can be observed that most edited videos have bigger \(b^e\) values than raw videos, which implies that in the training set, edited videos play leading roles in learning the property-weight of edited video summarization. Similarly, in Fig. \ref{Training_weight_for_raw}, the \(b^r\) values of raw videos are higher than edited videos, which indicates that in learning the property-weight of raw video summarization, the raw videos in the training set are more emphasized. Actually, the distribution of \(b^e\) and \(b^r\) meets our intuitions that in the learning procedure, inner-class videos should be more emphasized than inter-class videos.

\emph{Video features.} In the proposed framework, we use deep features to represent the video. Specifically, the VGGnet-16 \cite{simonyan2014very} network is employed to extract features for video frames, which is a popular convolutional neural network and has shown its effectiveness in several computer vision tasks. It has 13 convolutional layers and 3 fully connected layers. In practice, the frame feature is selected as the 4096 dimensional vector of the fc6 layer via the pre-trained model in \cite{DBLP:journals/corr/WangGH015}. Note that the feature of the shot is generated by averaging the feature vectors of its frames.

\emph{Learned property-weight.} It is worth mentioning that the learning procedure for the property-weight is executed for fifty times, each time with 30 randomly selected training videos (15 edited videos from TVsum50 and 15 raw videos from SumMe). Finally, the mean of the learning results are denoted as \(w^e\) and \(w^r\), which are depicted in Fig. \ref{Property_weight}. By comparing \(w^e\)\ and \(w^r\), we can clearly see the difference between edited and raw video summarization.

\subsection{Results on edited videos}\label{section4.2}

\begin{figure*}[t]
	\centering
	\includegraphics[width=\textwidth]{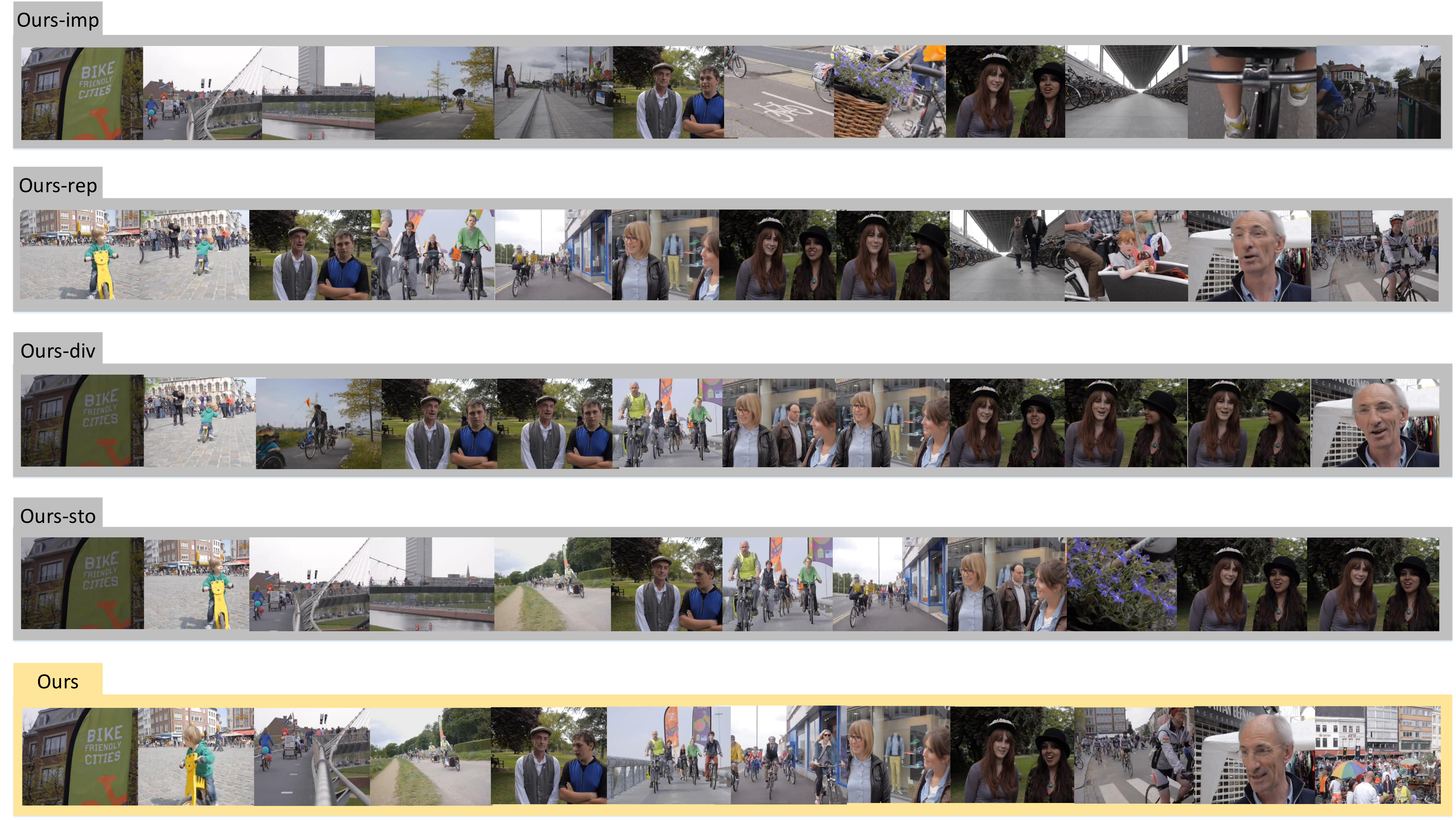}
	\caption{We can see from the summaries that this video is about a cycling affair, and some interviewees expressed their opinions on this affair. Carefully, we can see that the omission of any property model causes the decline of summary quality. For example, the summary generated by \(Ours-imp\) loses the focus on people, so some non-informative shots are selected. Besides, compared to the summary generated by our complete framework, we can also see the non-representative and redundant shots in the summaries generated by the other three versions.}\label{results_edited}
\end{figure*}

\textbf{Dataset}. The experiments about edited video summarization are carried on the TVsum50 dataset \cite{song2015tvsum}. This video dataset contains 50 edited videos, including news, documentary, etc. Their durations vary form 2-10 minutes. Each video is uniformly segmented into 2-second shots, and each shot is annotated with a user score, which is taken as the reference to generate the ground truth of video summary, i.e., high-score shots are selected as key-shots. Similar to existing approaches \cite{DBLP:conf/eccv/GygliGRG14,DBLP:conf/eccv/PotapovDHS14}, the length of the summary is constrained to less than 15\% of the video duration.

\textbf{Evaluation}. The quality of the automatically generated video summary is evaluated by its similarity with the ground truth. In this paper, the similarity is determined by the \(F\)-measure, which combines the scores of precision and recall for each pair of automatically generated summary and the ground truth. Note that two shots are matched only if they are exactly the same.

\begin{table}[h]
	\centering
	\caption{The results of various approaches on the TVsum50 dataset.}\label{Table1}
	\begin{tabular}{|c|c|c|}
		\hline
		Type&Approaches & F-measure \\
		\hline
		\hline
		Clustering  &VSUMM \cite{DBLP:journals/prl/AvilaLLA11} &0.383 \\
		\hline
		Dictionary learning&LiveLight \cite{DBLP:conf/cvpr/ZhaoX14a}&0.467  \\\hline
		Classifier&KVS \cite{DBLP:conf/eccv/PotapovDHS14}&0.471  \\\hline
		Domain knowledge&CA \cite{song2015tvsum}&0.513   \\
		\hline
		\hline
		
		Our framework&\textbf{Ours}&\textbf{0.527}   \\
		
		\hline

	\end{tabular}
	
\end{table}

\begin{table}[h]
	\centering
	\caption{The results of modified versions of our framework on the TVsum50 dataset.}\label{Table2}
	\begin{tabular}{|c|c|c|}
		\hline
		Elements in our framework&Versions & F-measure \\
		\hline
		\hline
		\multirow {4}*{Property models}  &Ours-imp &0.396 \\\cline{2-3}
		&Ours-rep&0.401   \\\cline{2-3}
		&Ours-div&0.434   \\\cline{2-3}
		&Ours-sto&0.513   \\\cline{2-3}
		\hline
		\hline
		\multirow {2}*{The combined training set}  &train-raw &0.483 \\\cline{2-3}
		&train-\({{b^e}}\)&0.451   \\\cline{2-3}
		
		\hline
		\hline
		Complete framework &\textbf{Ours}&\textbf{0.527}\\
		\hline
		
	\end{tabular}
	
\end{table}

\textbf{Results}. To verify the efficiency of the proposed framework, we compare the results with different types of edited video summarization approaches, including clustering \cite{DBLP:journals/prl/AvilaLLA11}, dictionary learning \cite{DBLP:conf/cvpr/ZhaoX14a}, classifier \cite{DBLP:conf/eccv/PotapovDHS14}, etc.. Table I illustrates the statistical results of the proposed framework and the comparing approaches.

To better analyze the results, the comparing approaches in Table \ref{Table1} are introduced briefly. VSUMM \cite{DBLP:journals/prl/AvilaLLA11} is based on \(k\)-means clustering algorithm. LiveLight \cite{DBLP:conf/cvpr/ZhaoX14a} aims to learn a compact dictionary from the original video as the video summary. KVS \cite{DBLP:conf/eccv/PotapovDHS14} is a supervised approach with an SVM classifier. It can be observed from Table \ref{Table1} that our framework achieves better performance than the above three comparing approaches. Additionally, CA \cite{song2015tvsum} is proposed with the dataset TVsum50. Note that CA summarizes the video with additional information, i.e., video title. However, our framework gets comparative performance even without this domain knowledge.
Generally, the results in Table \ref{Table1} indicate the efficiency of the general framework for edited video summarization.

To verify the necessity of the proposed four property models, we compare our framework with its modified versions, i.e., without one of the four property models. Table \ref{Table2} shows the results, where \(Ours-imp\) denotes our framework without the importance model, similar definitions is made for another 3 versions. It can be seen that omitting any property model from our framework will result in different reduction of the \(F\)-measure in some degree. Specifically, the reduction is positive correlated to their weights in our framework. In Fig. \ref{results_edited}, we exhibit one example of edited video summaries generated by our complete framework and its four modified versions. It can be seen that the summaries generated by the modified versions have more redundant, non-informative, or non-representative shots than the summary generated by our complete framework. The results illustrate that the four proposed property models are necessary for edited video summarization.

Additionally, to discuss the efficiency of the combined training set, we further compare the results with another two versions: 1) \(train-raw\) means that we just use edited videos for training. 2) \(train - {b^e}\) means the mixing-coefficient is ignored from the combined training set, i.e., the mixing-coefficients of all videos are 1. It can be seen from Table \ref{Table2} that the \(F\)-measures of the two versions are lower than our complete framework. Note that the \(F\)-measure of \(train - {b^e}\) is more lower than \(train-raw\). This is because that the rough combination of raw and edited videos causes structure mess in the training set. As a result, the interference of raw videos made for edited video summarization outweighs their contribution in the training procedure. After all, the decline of \(F\)-measure of these two versions indicates the efficiency of our combined training set with mixing-coefficient \({{b^e}}\).

\subsection{Results on short raw videos}\label{section4.3}

\begin{figure*}[t]
	\centering
	\includegraphics[width=\textwidth]{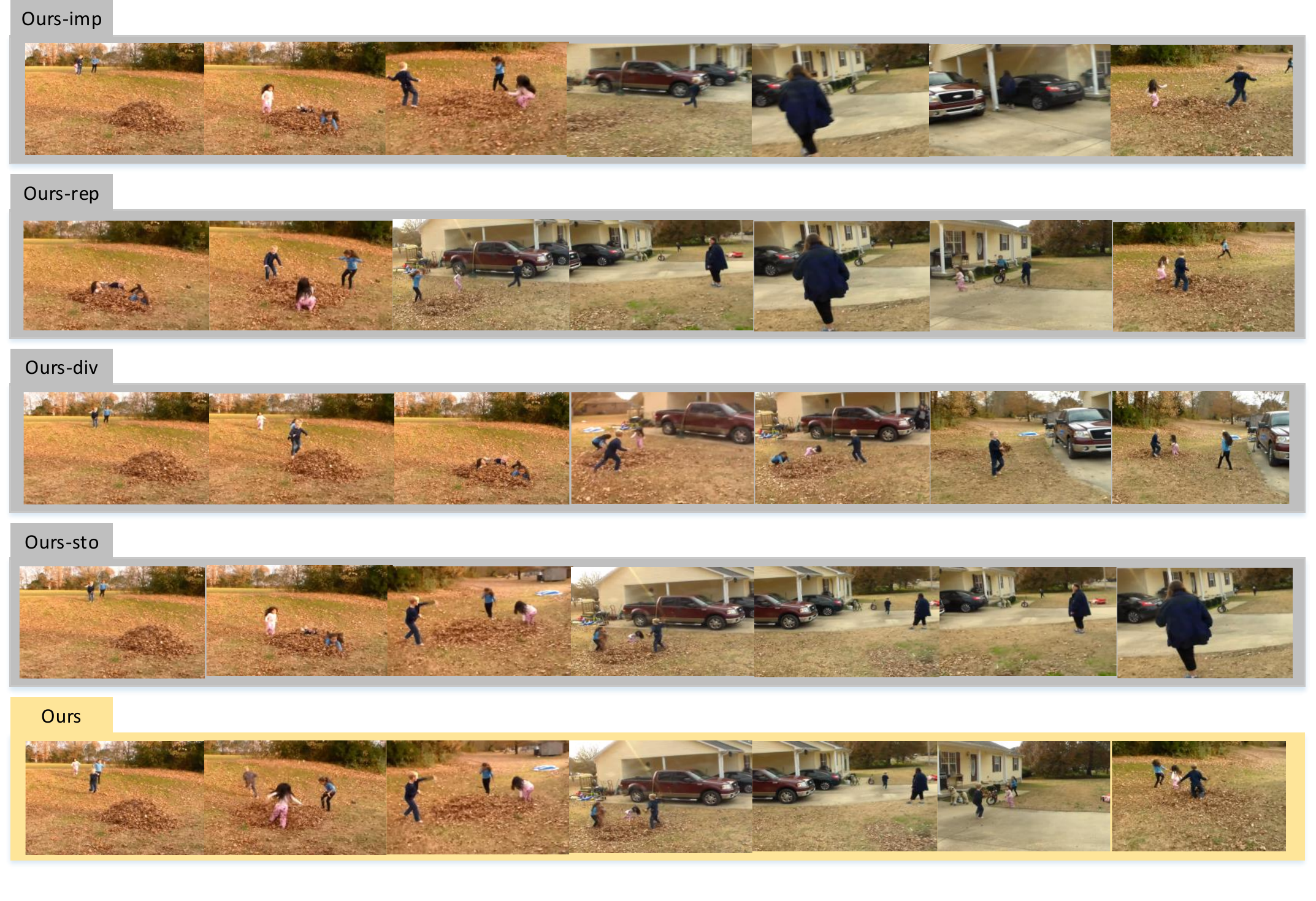}
	\caption{The video describes the kids playing in the leaves. The storyline has four stages: 1) Three kids run to a pile of leaves. 2) They play in the leaves. 3) Two kids go around to the back of the house. Another kid intercept them from the front of the house. 4) They return to the pile of leaves and play. By comparing the five summaries, we can see that the summary generated by our complete framework can capture the video content completely with the least non-informative, non-representative and redundant shots.}\label{fig_results_raw}
\end{figure*}

\textbf{Dataset.} The SumMe dataset is employed to verify the efficiency of our framework on raw video summarization. There are totally 25 videos, including holidays, events and sports. They are all raw videos, and the duration varies from 1 to 6 minutes. Each video is segmented into shots by the segmentation approach proposed in \cite{DBLP:conf/eccv/GygliGRG14}. Moreover, each video has about 15-18 human-made summaries as the ground truth. Their lengths vary from 5\% to 15\% of the video duration.

\textbf{Evaluation.} For each video, we compare the automatically generated summary with each human-made summary, and use the mean pairwise \(F\)-measure to quantify the summary quality. We need to highlight that when users generate a summary, the video are segmented according to their individual preference. While, when generating a summary automatically, the video is segmented by the segmentation approach in \cite{DBLP:conf/eccv/GygliGRG14}. Different segmentation patterns may cause that no two shots are exactly the same. Considering this, we calculate the overlap of the human-made summary and automatically-generated summary in frame-level.

\begin{table}[h]
	\centering
	\caption{The results of various approaches on the SumMe dataset.}\label{Table3}
	\begin{tabular}{|c|c|c|}
		\hline
		Category&Approaches & F-measure \\
		\hline
		\hline
		\multirow {3}*{Baselines}  &Random sampling &0.241 \\\cline{2-3}
		&Uniform sampling&0.273   \\\cline{2-3}
		&Clustering&0.356   \\\cline{2-3}
		
		\hline
		\hline
		Importance model &CSUV \cite{DBLP:conf/eccv/GygliGRG14}&0.393\\
		\hline
		\hline
		Supervised &LSMO \cite{gygli2015video}&0.397\\
		\hline
		\hline
		Our framework &\textbf{Ours}&\textbf{0.431}\\
		\hline
	\end{tabular}
	
\end{table}

\begin{table}[h]
	\centering
	\caption{The results of modified versions of our framework on the SumMe dataset.}\label{Table4}
	\begin{tabular}{|c|c|c|}
		\hline
		Elements in our framework&Versions & F-measure \\
		\hline
		\hline
		\multirow {4}*{Property models}  &Ours-imp &0.332 \\\cline{2-3}
		&Ours-rep&0.405   \\\cline{2-3}
		&Ours-div&0.411   \\\cline{2-3}
		&Ours-sto&0.413   \\\cline{2-3}
		\hline
		\hline
		\multirow {2}*{The combined training set}  &train-edited &0.390 \\\cline{2-3}
		&train-\({{b^r}}\)&0.372   \\\cline{2-3}
		
		\hline
		\hline
		Complete framework &\textbf{Ours}&\textbf{0.431}\\
		\hline
		
	\end{tabular}
	
\end{table}

\textbf{Results.} To evaluate the performance, we compare our framework with three baselines (i.e., random sampling, uniform sampling, clustering), and the state-of-the-art approaches, including CSUV \cite{DBLP:conf/eccv/GygliGRG14} and LSMO \cite{gygli2015video}. Table \ref{Table3} shows the experimental results. Where the clustering approach is based on \(k\)-means. CSUV \cite{DBLP:conf/eccv/GygliGRG14} is based on a single property model, interestingness, which has been introduced in the first part of Section \ref{section3.2}. LSMO is a supervised summarization approach \cite{gygli2015video}. We can see from Table \ref{Table3} that our framework achieves the best performance than other approaches. This shows the effectiveness of our framework for raw video summarization.

Then, experiments are executed to verify the necessity of the four property models and the efficiency of the combined training set. The results of the other modified versions of our framework are shown in Table \ref{Table4}. On one hand, it can be observed that removing any property model leads to the decline of the \(F\)-measure, which illustrates the necessity of the property models for raw video summarization. Fig. \ref{fig_results_raw} provides a detailed description about the influence of different property models to the summary quality. On the other hand, the decline of the results in \(train-edited\) and \(train-{b^r}\) shows that the combined training set is helpful with the mixing-coefficient \({{b^r}}\).

\subsection{Results on long raw videos}\label{section4.4}

\textbf{Dataset.} The ADL dataset contains 20 videos, most of them are about 30min long. Each of them is generated with a chest-mount GoPro camera, and records the wearers' daily life. During the shooting, the wearer is asked to do some daily activities, like combing hair, brushing teeth, etc.. Generally, there are totally 32 kinds of actions in the dataset, and every video contains about 18 kinds of actions, which are temporally annotated in the video. Simply, we segment the video into shots for the length of every 3 seconds, since it is enough for people to recognize the activities.

Note that the dataset is originally built for daily living activity detection in \cite{pirsiavash2012detecting}. But they are suitable for summarization since there are so many redundant and meaningless shots. Additionally, the summarization for daily living videos can help people get a more compact video as the diary to record their daily life.

\textbf{Evaluation.} The videos in the ADL dataset are so long that people share little agreement on the summarization. That's to say, it is hard to generate a unified human-made summary as the ground truth. The evaluation method by comparing automatically generated summary and the ground truth won't work.

Practically, the daily activities appearing in the video is the most important information to record the wearers' daily life. Inspired by this, we decide to use the percentage between the number of activities captured in the summary and occurring in the original video as the evaluation metric, denoted as \(Recall\) in the following equation.

\begin{equation}Recall = \frac{{\# activities\;in\;summary}}{{\# activities\;in\;video}}.\end{equation}

It should be mentioned that:

1)	Considering the same activities appearing in different video parts reflect different information, they are repeatedly counted in the evaluation procedure. As depicted in Fig. \ref{error}, the wearer uses computer twice in the video, i.e., 00:01-01:06 and 05:32-06:58. Carefully, we can see the two computers are different, indicating that the wearer have different purposes in the two activities. Therefore, they are double counted, so does the activity of drinking coffee/tea.

\begin{figure}[h] 
	\centering
	\includegraphics[width=0.4\textwidth]{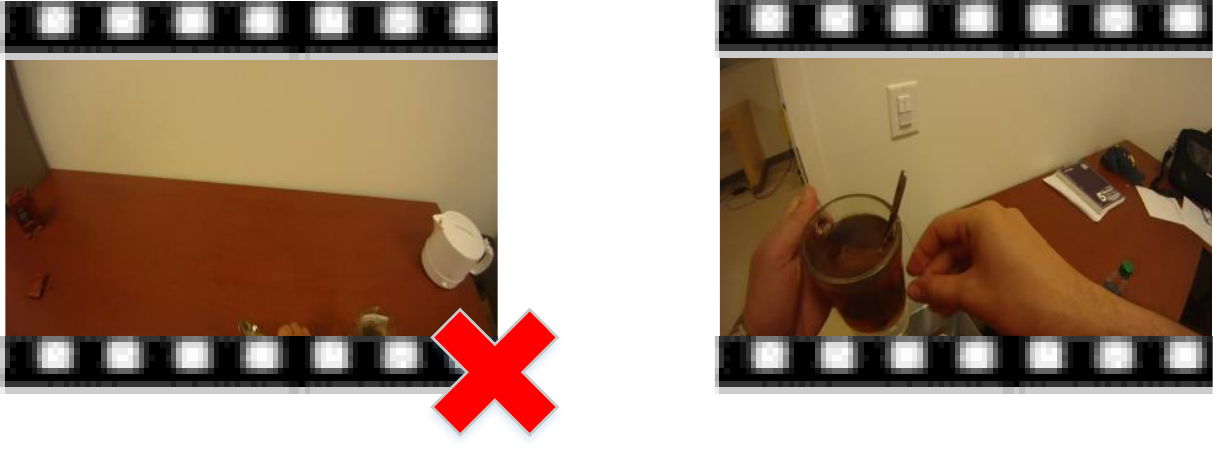} 
	\caption{The two shot are both labeled with "making tea", but only the right shot can be recognized.}\label{recognized}
\end{figure}

\begin{figure*}[t] 
	\centering
	\includegraphics[width=\textwidth]{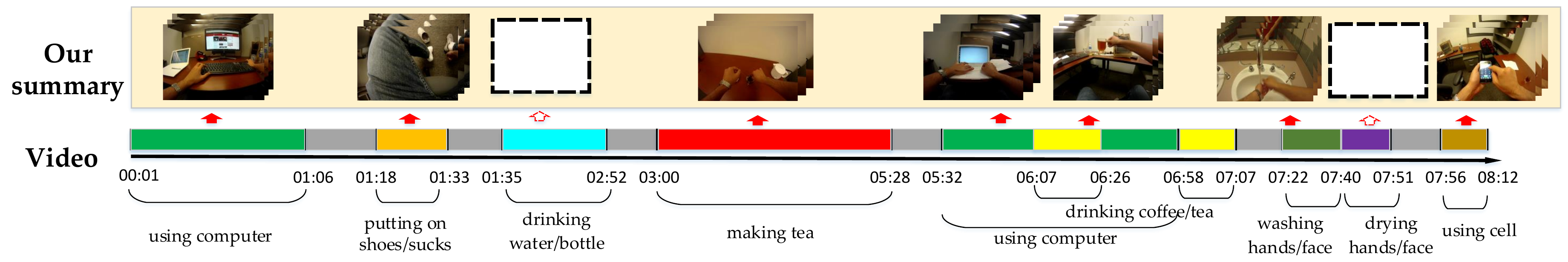} 
	\caption{It is an example about the summary of a long raw video, which is generated by our framework. The video is annotated with eleven activities, including using computer twice and drinking coffee/tea twice. These activities are highlighted in different colors, and gray means no activity labels. The two white boxes with dotted line stand for the activities missed in our summary, i.e., drinking water/bottle and drying hands/face. We can see that the key shots in our summary are informative, whose activities can be easily recognized. Best viewed in color.}\label{error}
\end{figure*}

2)	Initially, the summarization for daily life videos is to provide a wearer-friendly way to retrieve their daily life. The key-shot in the summary should catch the activity information correctly, and be easy for people to recognize. As depicted in Fig. \ref{recognized}, even though the key-shot has been labeled with one activity (i.e., making tea) in the original video, it doesn't mean the activity is captured by our summary, unless it can be recognized by people. Considering this, we organize 20 people to watch each generated summary, and ask them to recognize the activities occurring in it. It should be noted that only the activities labeled in the original video are considered. Practically, the mean number of activities recognized by the 20 people is denoted as the activities captured in the summary. 

\begin{table}[h]
	\centering
	\caption{The results of various approaches on the ADL dataset.}\label{Table5}
	\begin{tabular}{|c|c|c|}
		\hline
		Category&Approaches & Recall \\
		\hline
		\hline
		\multirow {3}*{Baselines}  &Random sampling &0.334 \\\cline{2-3}
		&Uniform sampling&0.362   \\\cline{2-3}
		&Clustering&0.475   \\\cline{2-3}
		
		\hline
		\hline
		
		Our framework &\textbf{Ours}&\textbf{0.583}\\
		\hline
	\end{tabular}
	
\end{table}

After summarization, the summary is shorter than 10\% of duration of the original video.
Table \ref{Table5} shows the summarization results of our framework and baselines. It can be observed that our framework outperforms the three baselines, which means the summary generated by our framework can better capture the activities in original video, meanwhile, reduce the redundancy. So our framework is helpful to generate video diary to record people's daily life. To better understand the quality of our summary, an example is shown in Fig. \ref{error}. It can be illustrated that most of the activities in the original video are included in our summary, which can be recognized easily from the key shots.

\subsection{Discussion} \label{section4.5}

 The summarization results on edited videos, short raw videos and long raw videos have shown the effectiveness of our framework. It benefits from:
 
 1) The four property models. Experimental results in Table \ref{Table2}, \ref{Table4} and Fig.\ref{results_edited}, \ref{fig_results_raw} have shown that the four property models, i.e., importance, representativeness, diversity and storyness, are necessary to both edited video and raw video summarization.
 
 2) The property-weights \(w^e\) and \(w^r\). The results in Fig. \ref{Property_weight} have proven that it is more reasonable to develop respective property-weight for edited video and raw video summarization.
 
 3) The mixing-coefficient \(\left( {{b^e},{b^r}} \right)\). The results in Table \ref{Table2} and \ref{Table4} indicate that the combined training set is helpful only if the mixing-coefficient is applied, since it can reduce the interference caused by inter-class training videos in combined training set.

Finally, it is worth mentioning that although the general framework is proposed for two classes of videos, i.e., edited videos and raw videos, it can also generalize to more classes, and the classification criteria is variable. It is easy to implement since the proposed property models in this paper are still applicable, and the only increment of computation is that more property-weights need to be learned (i.e., each class of videos own one property-weight) in the combined training set equipped with more mixing-coefficients.

\section{Conclusion} \label{section5}

In this paper, we have proposed a general summarization framework for edited videos and raw videos. The framework is a comprehensive score function with the weighted combination of four models, i.e., importance, representativeness, diversity and storyness, which are designed to capture the properties of the video summary. The weight of these models, denoted as property-weight, is learned in a supervised manner. Considering the characteristics of the two kinds of videos, we learn respective property-weight for both of them. Note that the training set is combined with both edited videos and raw videos in order to make up the lack of training data. Moreover, in the learning procedure, each training video is equipped with a pair of mixing-coefficient, which indicates its relevance to edited video and raw video summarization. The efficiency of our framework is verified on three datasets, including edited videos, short raw videos and long raw videos.

\bibliographystyle{IEEEtran}
\bibliography{TIP-16214-2016}

\end{document}